\documentclass[submission,publicdomain,creativecommons noncommercial]{eptcs}
\providecommand{\event}{STAI 2023} 

\usepackage{iftex}
\usepackage[decisionutilitycolor]{influence-diagrams}

\usepackage{booktabs}
\usepackage{amsmath}
\usepackage{amsthm}
\usepackage{algorithm}
\usepackage{algorithmic}
\usepackage{todonotes}
\usepackage{amsfonts}
\usepackage{amssymb}
\usepackage{wrapfig}
\usepackage{cleveref}
\usepackage{bm}
\usepackage{caption}
\usepackage{subcaption}

\ifpdf
  \usepackage{underscore}         
  \usepackage[T1]{fontenc}        
\else
  \usepackage{breakurl}           
\fi

\title{Experiments with Detecting and Mitigating AI Deception} 
\author{Ismail Sahbane
\institute{EPFL \\ Lausanne, Switzerland}
\email{ismail.sahbane@epfl.ch}
\and
Francis Rhys Ward
\institute{Imperial College London \\ London, UK}
\email{francis.ward19@imperial.ac.uk}
\and
C Henrik Åslund
\institute{Imperial College London \\ London, UK}
\email{c.aslund19@imperial.ac.uk}
}

\def\titlerunning{Detecting and mitigating deception}
\def\authorrunning{I. Sahbane \& F.R. Ward \& C.H. Åslund}

\begin{document}
\maketitle

\begin{abstract}
How to detect and mitigate deceptive AI systems is an open problem for the field of safe and trustworthy AI. We analyse two algorithms for mitigating deception:
The first is based on the path-specific objectives framework where paths in the game that incentivise deception are removed. 
The second is based on shielding, i.e., monitoring for unsafe policies and replacing them with a safe reference policy.
We construct two simple games and evaluate our algorithms empirically.
We find that both methods ensure that our agent is not deceptive, however, shielding tends to achieve higher reward.

\end{abstract}

\section{Introduction}

Deception is a challenge for building safe and trustworthy AI \cite{me}. Recent advances in reinforcement learning (RL) and language models (LMs) mean that we are increasingly living in a world containing highly capable, goal-directed \emph{agents} \cite{gpt4technicalreport2023}. Deception may be learned as an effective strategy for achieving goals in many environments, especially in multi-agent settings comprised of humans and AI agents \cite{me}. 

Technical work on deception in AI systems, and how to avoid it, is limited.
Deception has been defined within structural causal games (SCGs), which is a framework that applies causal graphs to game theory \cite{me}.
Given a causal graph, it is possible to ensure certain safety properties by removing paths in that causal graph \cite{Farquhar2022Jun}. Giving learning agents these kinds of \emph{path-specific objectives} (PSOs) is also applicable when deception is the property that is considered unsafe. These methods ensure that the agent will not deceive a human providing feedback, however, it will also ensure that the agent will not persuade, teach, or coordinate with the human (i.e., it will not try to influence the human in any way).
Shielding is another class of methods for ensuring safety in learning agents \cite{OdriozolaOlaldeZA23, carr2023safe, waga2022dynamic, KonighoferBEP22, Elsayed-AlyBAET21, giacobbe2021shielding, AlshiekhBEKNT18}. A shield is a kind of monitor. In addition to monitoring, the shield replaces an unsafe action with a safe action if the verification returns that the policy does not satisfy the safety specification.

In this paper, we make three contributions:
(1)~We introduce a shielding ~algorithm to ensure that an agent does not deceive other agents.
(2)~We introduce two simple games to evaluate deception in agents.
(3)~We evaluate our algorithm in these environments against an algorithm based on PSO.
This paper is organised as follows: In Section~\ref{sec:DefiningAndDetectingDeception}, we recapitulate the definition of deception in SCGs. In Section~\ref{sec:MitigatingDeception}, we introduce our algorithm and compare it to the PSO~algorithm .
We then conclude in Section~\ref{sec:Conclusion}.

\section{Defining and detecting deception}
\label{sec:DefiningAndDetectingDeception}

\textbf{Structural Causal Games (SCGs)} offer a representation of causality in games \cite{Hammond}. 
An SCG is a directed acyclic graph containing variables and causal edges between them. There are three types of variables: chance variables ($X$), decision variables ($D$) and utility variables ($U$). Along with the graph, an SCG defines the conditional probability distribution (CPD) over each (non-decision) variable, given its parents. Agent's policies choose the CPD over decision variables and agents choose their policies to maximise the expected sum of utility and we use the Nash equilibrium concept. At the beginning of the game a \emph{setting} is sampled from the prior over the game; given a setting and a policy profile, the value of any variable is uniquely determined.
 Kenton et al. \cite{kenton2022discovering} define \emph{agents} in SCGs as systems that would adapt their policy if their actions influenced the world in a different way.
This is the relevant notion of agency, as
we define belief and intent based on how the agent would adapt its behaviour to such changes. 

We now introduce a simple signalling game, where an agent that can be weak or strong tries to avoid being attacked by another agent, that wants to attack them only if they're weak. They can defend or retreat, and their decision is observed by the other agent.

\begin{wrapfigure}{L}{0.3\textwidth}
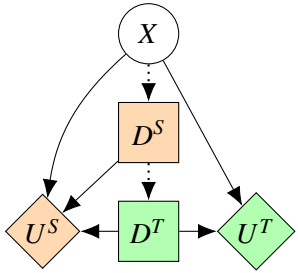

          \centering
          \vspace{-2mm}
\begin{influence-diagram}

  \node (DS) [decision, player1] {$D^{S}$};
  \node(DT) [below =0.5 of DS, decision, player2] {$D^{T}$};
  \node (X) [above =0.5 of DS] {$X$};
  \node (US) [left =0.5 of DT, utility, player1] {$U^S$};
  \node (UT) [right =0.5 of DT, utility, player2] {$U^{T}$};

  \edge[information] {X} {DS};
  \edge[information] {DS} {DT};
  \edge {DS} {US};
  \edge {DT} {UT};
  \edge {DT} {US};
\path (X) edge[ ->, bend right=20] (US);
\edge {X} {UT};
\end{influence-diagram} 
\caption{Ex.1 SCG graph. Chance variables are circular, decisions square, utilities diamond and the latter two are colour coded by their association with different agents. Solid edges represent causal dependence and dotted edges are information links.}
         \label{fig:run}
         \vspace{-4mm} 
\end{wrapfigure}

\textbf{Example 1 (War game \cref{fig:run}):} A signaller S has type $X \in \{strong, weak\}$. At the start of the game, $S$ observes $X$, but the target agent $T$ does not. The agents have decisions $D^S \in \{retreat, defend\}$
and $D^T \in \{\neg attack, attack\}$. A weak $S$ prefers to retreat whereas a strong $S$ prefers to defend. $T$ prefers to attack only if $S$ is weak. Regardless of type, $S$ does not want to be attacked (and cares more about being attacked than about their own action). $X$ follows a $Bernoulli(0.9)$ distribution so that $S$ is strong with probability $0.9$. $U^T = 1$ if $T$ attacks a weak $S$ or does not attack a strong $S$, and 0 otherwise. $S$ gains utility $2$ for not getting attacked, and utility $1$ for performing the action preferred by their type (e.g., utility $1$ for retreating if they are weak).

 \emph{To deceive is to intentionally cause to have a false belief that is not believed to be true} \cite{mahon}. Past work defines belief, intention, and deception in SCGs \cite{me}; these definitions 
 only refer to agent behaviour.
 
\textbf{Belief}
Agents have beliefs over \emph{propositions} $\phi$, i.e., Boolean formula of variable assignments (e.g., $\phi: X=x \land \neg Y=y$).
An agent \emph{believes} a proposition $\phi$ if 1) they act as though they observed $\phi$ is true; 2) they would have acted differently if they observed $\phi$ was false.
An agent has a \emph{true/false belief} if they believe $\phi$ and $\phi = true/false$. 

\textbf{Example 1 (continued):} Since $S$'s probability of being weak is low, its optimal policy is to always defend, in order to signal a strong type. $T$'s best policy in this case is to attack if and only if $S$ retreats. These two policies form a Nash equilibrium. When $X = weak$, $T$ believes the proposition $\phi : X=Strong$, as 1) if they had observed that $X=Strong$, they would not have attacked, and 2) if they had observed that $X=Weak$, they would attack. Therefore, they respond to $\phi$, and they act as if $\phi=true$, so the two conditions for belief are met. When $X = weak$, $\phi = false$, so $T$ has a false belief about $\phi$.

\textbf{Intention}
Previous work defines notions of intention, suitable for algorithms, in causal models \cite{DBLP:conf/aaai/HalpernK18,ashton-intent,me}. Essentially, an agent \emph{intentionally causes} the outcomes which provide sufficient reason for it to choose its policy over an alternate policy. What \emph{intent to cause} means, intuitively, is that if the outcomes the agent wanted were guaranteed
to happen anyway, they would not mind choosing an alternative policy \cite{me}. \looseness=-1


\textbf{Example 1 (continued):} Under the Nash policy, $S$ intends to cause $T$ to not attack, w.r.t.\ the alternative (honest) policy ($defend$ when $X=strong$ and $retreat$ when $X=weak$), because if $T$'s policy was fixed to $\neg attack$, then  $S$ would choose the honest policy. 
Hence, $S$ intends to cause $D^T = \neg attack$.

\textbf{Deception}
An agent $S$ \emph{deceives} an agent $T$ about a proposition $\phi$ if 1) $S$ intentionally causes $T$'s decision $D^T$; 2) $T$ has a false belief about $\phi$; 3) $S$ does not believe $\phi$.

\textbf{Example 1 (continued):} Under the Nash policies, $S$ deceives $T$ about their type when $S$ is $weak$. As seen above, $T$ believes that $X=Strong$, $S$ intends for $D^T$ to be $\neg attack$, and $S$ does not believe that $X=weak$, 
so all the conditions for deception are met.

  \section{Mitigating Deception}
  \label{sec:MitigatingDeception}

We perform experiments in two examples, optimising agents to play these two games with no mitigation, with PSO, and with shielding. We compare these methods on their optimality and deceptiveness. 

\begin{wraptable}{R}{0.6\textwidth}
\vspace{-2mm}
\hrule
\captionof{algorithm}{\textsc{Path-Specific Objectives}}
\vspace{-3mm}
\hrule
 \label{alg:pso}
 \textbf{Input}: An SCG $\mathcal{M} = (\mathcal{G}, \theta)$, graphical criterion $\mathcal{C}$, policies $\pi^{-i}$, natural distributions $\mathcal{N}$. \\
\textbf{Output}: PSO-optimal policy $\pi^i$
\begin{algorithmic}[1] 
\STATE Reduce $\mathcal{G}$ to $\mathcal{G}'$ using $\mathcal{C}$. 
\STATE Impute policies $\pi^{-i}$ and natural distributions from $\mathcal{N}$ to those variables with fewer parents in $\mathcal{G}'$ to obtain $\theta'$.
\STATE Train an agent in $\mathcal{M}' = (\mathcal{G}', \theta')$ to obtain policy $\pi^i$.
\end{algorithmic}
\hrule
\end{wraptable}

\textbf{Path Specific Objective (PSO) \cite{Farquhar2022Jun}}
prevents $S$ learning a deceptive policy by removing $S$'s ability to influence $T$'s decision during training. In SGSs, this corresponds to removing the path in the graph between $D^S$ and $D^T$. 
The PSO algorithm is shown in \cref{alg:pso}.

\textbf{Example 1 (continued):} In the war game, we remove the edge from $D^S$ to $D^T$. $S$ is now only interested by the utility it gets directly from its decision, meaning it will learn the honest policy. Therefore, $S$ can be trained with PSO, and it will play the game without being deceptive. 
However, this removes $S$'s ability to learn to influence $T$ in any way, including positively. In the following example, the only strategy for achieving utility is to influence the other agent, and so the PSO agent does not learn anything. 

\begin{wrapfigure}{L}{0.3\textwidth}
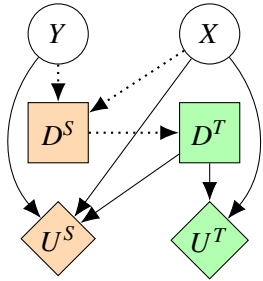

    \centering
    \vspace{-4mm}
\begin{influence-diagram}

      \node (X) {$X$};
      \node (Y) [left =1.2 of X] {$Y$};
      \node (DS) [below =0.5 of Y, decision, player1] {$D^S$};
      \node (DT) [below =0.5 of X, decision, player2] {$D^T$};
      \node (US) [below =0.5 of DS, utility, player1] {$U^S$};
      \node (UT) [below =0.5 of DT, utility, player2] {$U^T$};
      
      \edge[information] {X} {DS};
      \edge[information] {Y} {DS};
      \edge[information] {DS} {DT};
      \edge {DT} {US};
      \path (X) edge[ ->] (US);
      \path (Y) edge[ ->, bend right = 40] (US);
      \edge {DT} {UT};
      \path (X) edge[ ->, bend left = 40] (UT);
      
\end{influence-diagram}
\caption{Ex. 2 SCG Graph}
    \label{fig:2}
    \vspace{-4mm}
\end{wrapfigure}

\textbf{Example 2 (\cref{fig:2}):} Variables $X \in \{0, 1, 2\}$ is uniformly sampled, and $Y \in \{0, 1\}$ follows a $Ber(0.1)$ distribution. $S$ observes both $X$ and $Y$, while $T$ does not. $D^T \in \{0,1,2\}$ and $T$'s objective is to correctly bet on the value of $X$. $D^S \in \{0,1,2\}$, and $S$ gets some utility when $T$ correctly bets on $X$, but when $Y=1$, $S$ can get more utility if $D^T = (X+1)\mod{3}$. $T$ observes $D^S$.
\\ This example differs from the first one as $S$'s only way to get utility is to influence $T$'s decision. $S$ can adopt two sensible policies, which are to always report $X$ (the honest policy), or to, when $Y=1$, report on $(X+1)\mod{3}$ instead (a deceptive policy, $S$ intentionally causes $T$ to believe that $X$ has the wrong value). Since $Y$ is rarely $1$, as before, $T$'s optimal strategy remains to follow $S$'s decision even if it is sometimes deceptive. 
If we use PSO with this game, $S$ will have no way to influence its utility, and $T$ will have no information on the value of $X$. Therefore, any policies for $S$ and $T$ are optimal and the agents will not learn anything. We introduce the shielding algorithm to solve this problem by preventing deceptive policies from being learned, in a more fine-grained way than PSO. 

\begin{wraptable}{l}{0.65\textwidth}
\vspace{-2mm}
\hrule
\captionof{algorithm}{\textsc{Shielding}} \vspace{-3mm}
\hrule
 \textbf{Input}: SCG $\mathcal{M}=(\mathcal G, \theta)$, reference policy~$\pi_\mathrm{ref}$, $T$'s policy $\pi^T$, proposition $\phi$ 
 \\
\textbf{Output}: optimal non-deceptive policy $\pi$
\begin{algorithmic}[1] 
\STATE $\pi\leftarrow\pi_\mathrm{ref}$
\FOR{$s$ \textbf{in} settings}
\STATE take action $a$ sampled from $\pi(\cdot|s)$
\STATE calculate optimal policy $\pi_\star$
\STATE D$\leftarrow$\textsc{DeceptionCheck}($\mathcal{M}$, ~$\bm{\pi}$=$ (\pi_\star, \pi^T)$, $\pi_\mathrm{ref}$=$\pi$, $S$, $T$, $\phi$)
\STATE \textbf{if} D=false \textbf{then} $\pi\leftarrow\pi_\star$
\ENDFOR
\end{algorithmic}
\hrule
\vspace{-4mm}
\end{wraptable}

\textbf{Shielding }
 uses a safety constraint, and checks that an agent's action or policy satisfies the constraint before letting it perform that action. In our case, we shield the whole policy rather than individual decisions, and we use the shield during training to prevent the agent learning deceptive policies. The shield used is  the deception verification 
presented in \Cref{alg:deceptioncheck}.

We built a simple environment to apply the above definitions and to investigate ways to train non-deceptive agents in SCGs. 
$S$~(the potentially deceptive agent) is being trained with a minimal version of RL, where $S$~plays several games with random settings and policy. 
We assume that $T$~(the target of deception) has fixed Nash-policy, because we work with games where the occasions for $S$ to benefit from deception are rare, and the best policy for~$T$ remains to believe $S$'s~signal, despite it being sometimes false.
We implemented \cref{alg:deceptioncheck}, which, given a policy, and a reference policy, indicates whether the policy is deceptive w.r.t the reference policy, by testing every possible setting. Hence \cref{alg:deceptioncheck} is complete and sound. 
We initialize a known-safe policy  as the reference. As soon as a better performing safe policy is found, we use this as the new reference. \looseness=-1
\begin{algorithm}
\caption{\textsc{Deception Check}}
\label{alg:deceptioncheck}
\textbf{Input}: SCG $\mathcal{M}=(\mathcal G, \theta)$, policy profile~$\bm{\pi}$, referencepolicy~$\pi_\mathrm{ref}$, agent~$S$, agent~$T$, proposition~$\phi$ \\
\textbf{Output}: boolean, true if $S$ deceives $T$ about $\phi$ under $\bm{\pi}$ w.r.t. $\pi_\mathrm{ref}$
\begin{algorithmic}[1]
\FOR{$s$ \textbf{in} settings}
\STATE initialise $\mathcal{M}$ with $s$ and $\bm{\pi}$ 
\STATE compute whether $S$ intends to cause $D^T$ under $\mathcal{M}$, $\bm{\pi}$ and $\pi_\mathrm{ref}$
\STATE compute whether $S$ and $T$ believe $\phi$ under $\mathcal{M}$, $\bm{\pi}$,
\STATE compute whether $\phi$ is true under $\mathcal{M}$ and $s$
\STATE deceptive $\leftarrow$ $S$ intends to cause $D^T$ $\And$ $T$ believes $\phi$ $\And$ $S$ does not believe $\phi$ $\And \phi$ is false
\STATE \textbf{if} deceptive \textbf{then return} true
\ENDFOR
\STATE \textbf{if not} deceptive \textbf{then return} false 
\end{algorithmic}
\end{algorithm}



\begin{wraptable}{R}{0.73\textwidth} 
\vspace{-6mm}
\centering
    \caption{Results for examples 1 and 2} \vspace{-3mm}
    \begin{tabular}{l c c c c}
        \hline
        & \multicolumn{2}{c}{ex. 1} & \multicolumn{2}{c}{ex. 2} \\
        & deceptive &  performance & deceptive & performance \\
        \hline
        shielding \textbf{(our)} & \textbf{no} & optimal-honest & \textbf{no} & optimal-honest \\
        PSO & \textbf{no} & optimal-honest & \textbf{no} & sub-optimal \\
        no mitigation & yes & \textbf{optimal} & yes & \textbf{optimal} \\
        \hline
    \end{tabular} \label{tab:results}
\end{wraptable} 


\emph{Results} are summarised in \cref{tab:results}. For Ex.1, both PSO and shielding learn the optimal non-deceptive policy, whereas when no mitigation is used the optimal (deceptive) policy is learned. For Ex.2, shielding learns the optimal non-deceptive policy, but the PSO-agent cannot learn anything, as in this example the only way for $S$ to gain utility is to influence $T$. 
\looseness=-1


\section{Conclusion}
\label{sec:Conclusion}
\emph{Summary} We introduce a novel shielding algorithm for mitigating deceptive learning agents. We show, in two toy environments, that our algorithm has advantages over previous methods. 

\emph{Limitations and future work} The examples are simplistic, and the optimal policies are very easy to find analytically without doing any training. 
This work acts as a proof of concept for the idea of automatically detecting and preventing deception while training. Many simplifying assumptions are made, e.g. the fact that the games only have one time-step, or the assumption that one of the policies is fixed. In addition, the verification is exhaustive on the setting space. This works with the small domains of these examples but might become intractable for larger and more realistic problems, which could require Monte-Carlo sampling of the setting, or a latent representation of it. Furthermore, shielding requires an initial safe reference policy, and its convergence to good safe policies is unknown.

\nocite{*}
\bibliographystyle{eptcs}
\bibliography{literature}

\begin{thebibliography}{10}
\providecommand{\bibitemdeclare}[2]{}
\providecommand{\surnamestart}{}
\providecommand{\surnameend}{}
\providecommand{\urlprefix}{Available at }
\providecommand{\url}[1]{\texttt{#1}}
\providecommand{\href}[2]{\texttt{#2}}
\providecommand{\urlalt}[2]{\href{#1}{#2}}
\providecommand{\doi}[1]{doi:\urlalt{https://doi.org/#1}{#1}}
\providecommand{\eprint}[1]{arXiv:\urlalt{https://arxiv.org/abs/#1}{#1}}
\providecommand{\bibinfo}[2]{#2}

\bibitemdeclare{article}{Farquhar2022Jun}
\bibitem{Farquhar2022Jun}
\bibinfo{author}{Sebastian \surnamestart Farquhar~et al.\surnameend}
  (\bibinfo{year}{2022}): \emph{\bibinfo{title}{{Path-Specific Objectives for
  Safer Agent Incentives}}}.
\newblock {\slshape \bibinfo{journal}{AAAI}}
  \bibinfo{volume}{36}(\bibinfo{number}{9}), pp. \bibinfo{pages}{9529--9538},
  \doi{10.1609/aaai.v36i9.21186}.

\bibitemdeclare{inproceedings}{AlshiekhBEKNT18}
\bibitem{AlshiekhBEKNT18}
\bibinfo{author}{Mohammed \surnamestart Alshiekh\surnameend},
  \bibinfo{author}{Roderick \surnamestart Bloem\surnameend},
  \bibinfo{author}{R{\"{u}}diger \surnamestart Ehlers\surnameend},
  \bibinfo{author}{Bettina \surnamestart K{\"{o}}nighofer\surnameend},
  \bibinfo{author}{Scott \surnamestart Niekum\surnameend} \&
  \bibinfo{author}{Ufuk \surnamestart Topcu\surnameend} (\bibinfo{year}{2018}):
  \emph{\bibinfo{title}{Safe Reinforcement Learning via Shielding}}.
\newblock In \bibinfo{editor}{Sheila~A. \surnamestart McIlraith\surnameend} \&
  \bibinfo{editor}{Kilian~Q. \surnamestart Weinberger\surnameend}, editors:
  {\slshape \bibinfo{booktitle}{Proceedings of the Thirty-Second {AAAI}
  Conference on Artificial Intelligence, (AAAI-18), the 30th innovative
  Applications of Artificial Intelligence (IAAI-18), and the 8th {AAAI}
  Symposium on Educational Advances in Artificial Intelligence (EAAI-18), New
  Orleans, Louisiana, USA, February 2-7, 2018}}, \bibinfo{publisher}{{AAAI}
  Press}, pp. \bibinfo{pages}{2669--2678}.
\newblock
  \urlprefix\url{https://www.aaai.org/ocs/index.php/AAAI/AAAI18/paper/view/17211}.

\bibitemdeclare{article}{ashton-intent}
\bibitem{ashton-intent}
\bibinfo{author}{Hal \surnamestart Ashton\surnameend} (\bibinfo{year}{2022}):
  \emph{\bibinfo{title}{Definitions of intent suitable for algorithms}}.
\newblock {\slshape \bibinfo{journal}{Artificial Intelligence and Law}}, pp.
  \bibinfo{pages}{1--32}.

\bibitemdeclare{book}{bostrom2017superintelligence}
\bibitem{bostrom2017superintelligence}
\bibinfo{author}{Nick \surnamestart Bostrom\surnameend} (\bibinfo{year}{2017}):
  \emph{\bibinfo{title}{Superintelligence}}.
\newblock \bibinfo{publisher}{Dunod}.

\bibitemdeclare{article}{carlsmith2022}
\bibitem{carlsmith2022}
\bibinfo{author}{Joseph \surnamestart Carlsmith\surnameend}
  (\bibinfo{year}{2022}): \emph{\bibinfo{title}{Is Power-Seeking {AI} an
  Existential Risk?}}
\newblock {\slshape \bibinfo{journal}{CoRR}} \bibinfo{volume}{abs/2206.13353},
  \doi{10.48550/arXiv.2206.13353}.
\newblock \eprint{2206.13353}.

\bibitemdeclare{inproceedings}{carr2023safe}
\bibitem{carr2023safe}
\bibinfo{author}{Steven \surnamestart Carr\surnameend}, \bibinfo{author}{Nils
  \surnamestart Jansen\surnameend}, \bibinfo{author}{Sebastian \surnamestart
  Junges\surnameend} \& \bibinfo{author}{Ufuk \surnamestart Topcu\surnameend}
  (\bibinfo{year}{2023}): \emph{\bibinfo{title}{Safe reinforcement learning via
  shielding under partial observability}}.
\newblock In: {\slshape \bibinfo{booktitle}{AAAI}}.

\bibitemdeclare{inproceedings}{Elsayed-AlyBAET21}
\bibitem{Elsayed-AlyBAET21}
\bibinfo{author}{Ingy \surnamestart Elsayed{-}Aly\surnameend},
  \bibinfo{author}{Suda \surnamestart Bharadwaj\surnameend},
  \bibinfo{author}{Christopher \surnamestart Amato\surnameend},
  \bibinfo{author}{R{\"{u}}diger \surnamestart Ehlers\surnameend},
  \bibinfo{author}{Ufuk \surnamestart Topcu\surnameend} \&
  \bibinfo{author}{Lu~\surnamestart Feng\surnameend} (\bibinfo{year}{2021}):
  \emph{\bibinfo{title}{Safe Multi-Agent Reinforcement Learning via
  Shielding}}.
\newblock In \bibinfo{editor}{Frank \surnamestart Dignum\surnameend},
  \bibinfo{editor}{Alessio \surnamestart Lomuscio\surnameend},
  \bibinfo{editor}{Ulle \surnamestart Endriss\surnameend} \&
  \bibinfo{editor}{Ann \surnamestart Now{\'{e}}\surnameend}, editors: {\slshape
  \bibinfo{booktitle}{{AAMAS} '21: 20th International Conference on Autonomous
  Agents and Multiagent Systems, Virtual Event, United Kingdom, May 3-7,
  2021}}, \bibinfo{publisher}{{ACM}}, pp. \bibinfo{pages}{483--491},
  \doi{10.5555/3463952.3464013}.
\newblock
  \urlprefix\url{https://www.ifaamas.org/Proceedings/aamas2021/pdfs/p483.pdf}.

\bibitemdeclare{article}{giacobbe2021shielding}
\bibitem{giacobbe2021shielding}
\bibinfo{author}{Mirco \surnamestart Giacobbe\surnameend},
  \bibinfo{author}{Mohammadhosein \surnamestart Hasanbeig\surnameend},
  \bibinfo{author}{Daniel \surnamestart Kroening\surnameend} \&
  \bibinfo{author}{Hjalmar \surnamestart Wijk\surnameend}
  (\bibinfo{year}{2021}): \emph{\bibinfo{title}{Shielding atari games with
  bounded prescience}}.
\newblock {\slshape \bibinfo{journal}{arXiv preprint arXiv:2101.08153}}.

\bibitemdeclare{inproceedings}{DBLP:conf/aaai/HalpernK18}
\bibitem{DBLP:conf/aaai/HalpernK18}
\bibinfo{author}{Joseph~Y. \surnamestart Halpern\surnameend} \&
  \bibinfo{author}{Max \surnamestart Kleiman{-}Weiner\surnameend}
  (\bibinfo{year}{2018}): \emph{\bibinfo{title}{Towards Formal Definitions of
  Blameworthiness, Intention, and Moral Responsibility}}.
\newblock In \bibinfo{editor}{Sheila~A. \surnamestart McIlraith\surnameend} \&
  \bibinfo{editor}{Kilian~Q. \surnamestart Weinberger\surnameend}, editors:
  {\slshape \bibinfo{booktitle}{Proceedings of the Thirty-Second {AAAI}
  Conference on Artificial Intelligence, (AAAI-18), the 30th innovative
  Applications of Artificial Intelligence (IAAI-18), and the 8th {AAAI}
  Symposium on Educational Advances in Artificial Intelligence (EAAI-18), New
  Orleans, Louisiana, USA, February 2-7, 2018}}, \bibinfo{publisher}{{AAAI}
  Press}, pp. \bibinfo{pages}{1853--1860}.
\newblock
  \urlprefix\url{https://www.aaai.org/ocs/index.php/AAAI/AAAI18/paper/view/16824}.

\bibitemdeclare{article}{Hammond}
\bibitem{Hammond}
\bibinfo{author}{Lewis \surnamestart Hammond\surnameend},
  \bibinfo{author}{James \surnamestart Fox\surnameend}, \bibinfo{author}{Tom
  \surnamestart Everitt\surnameend}, \bibinfo{author}{Ryan \surnamestart
  Carey\surnameend}, \bibinfo{author}{Alessandro \surnamestart
  Abate\surnameend} \& \bibinfo{author}{Michael \surnamestart
  Wooldridge\surnameend} (\bibinfo{year}{2023}):
  \emph{\bibinfo{title}{Reasoning about causality in games}}.
\newblock {\slshape \bibinfo{journal}{Artificial Intelligence}}
  \bibinfo{volume}{320}, p. \bibinfo{pages}{103919},
  \doi{10.1016/j.artint.2023.103919}.

\bibitemdeclare{article}{kenton2022discovering}
\bibitem{kenton2022discovering}
\bibinfo{author}{Zachary \surnamestart Kenton\surnameend},
  \bibinfo{author}{Ramana \surnamestart Kumar\surnameend},
  \bibinfo{author}{Sebastian \surnamestart Farquhar\surnameend},
  \bibinfo{author}{Jonathan \surnamestart Richens\surnameend},
  \bibinfo{author}{Matt \surnamestart MacDermott\surnameend} \&
  \bibinfo{author}{Tom \surnamestart Everitt\surnameend}
  (\bibinfo{year}{2022}): \emph{\bibinfo{title}{Discovering Agents}}.
\newblock {\slshape \bibinfo{journal}{arXiv preprint arXiv:2208.08345}}.

\bibitemdeclare{inproceedings}{KonighoferBEP22}
\bibitem{KonighoferBEP22}
\bibinfo{author}{Bettina \surnamestart K{\"{o}}nighofer\surnameend},
  \bibinfo{author}{Roderick \surnamestart Bloem\surnameend},
  \bibinfo{author}{R{\"{u}}diger \surnamestart Ehlers\surnameend} \&
  \bibinfo{author}{Christian \surnamestart Pek\surnameend}
  (\bibinfo{year}{2022}): \emph{\bibinfo{title}{Correct-by-Construction Runtime
  Enforcement in {AI} - {A} Survey}}.
\newblock In \bibinfo{editor}{Jean{-}Fran{\c{c}}ois \surnamestart
  Raskin\surnameend}, \bibinfo{editor}{Krishnendu \surnamestart
  Chatterjee\surnameend}, \bibinfo{editor}{Laurent \surnamestart
  Doyen\surnameend} \& \bibinfo{editor}{Rupak \surnamestart
  Majumdar\surnameend}, editors: {\slshape \bibinfo{booktitle}{Principles of
  Systems Design - Essays Dedicated to Thomas A. Henzinger on the Occasion of
  His 60th Birthday}}, {\slshape \bibinfo{series}{Lecture Notes in Computer
  Science}} \bibinfo{volume}{13660}, \bibinfo{publisher}{Springer}, pp.
  \bibinfo{pages}{650--663}, \doi{10.1007/978-3-031-22337-2\_31}.
\newblock \urlprefix\url{https://doi.org/10.1007/978-3-031-22337-2\_31}.

\bibitemdeclare{incollection}{mahon}
\bibitem{mahon}
\bibinfo{author}{James~Edwin \surnamestart Mahon\surnameend}
  (\bibinfo{year}{2016}): \emph{\bibinfo{title}{{The Definition of Lying and
  Deception}}}.
\newblock In \bibinfo{editor}{Edward~N. \surnamestart Zalta\surnameend},
  editor: {\slshape \bibinfo{booktitle}{The {Stanford} Encyclopedia of
  Philosophy}}, \bibinfo{edition}{{W}inter 2016} edition,
  \bibinfo{publisher}{Metaphysics Research Lab, Stanford University}.

\bibitemdeclare{inproceedings}{OdriozolaOlaldeZA23}
\bibitem{OdriozolaOlaldeZA23}
\bibinfo{author}{Haritz \surnamestart Odriozola{-}Olalde\surnameend},
  \bibinfo{author}{Maider \surnamestart Zamalloa\surnameend} \&
  \bibinfo{author}{Nestor \surnamestart Arana{-}Arexolaleiba\surnameend}
  (\bibinfo{year}{2023}): \emph{\bibinfo{title}{Shielded Reinforcement
  Learning: {A} review of reactive methods for safe learning}}.
\newblock In: {\slshape \bibinfo{booktitle}{{IEEE/SICE} International Symposium
  on System Integration, {SII} 2023, Atlanta, GA, USA, January 17-20, 2023}},
  \bibinfo{publisher}{{IEEE}}, pp. \bibinfo{pages}{1--8},
  \doi{10.1109/SII55687.2023.10039301}.
\newblock \urlprefix\url{https://doi.org/10.1109/SII55687.2023.10039301}.

\bibitemdeclare{article}{gpt4technicalreport2023}
\bibitem{gpt4technicalreport2023}
\bibinfo{author}{\surnamestart OpenAI\surnameend} (\bibinfo{year}{2023}):
  \emph{\bibinfo{title}{{GPT-4} Technical Report}}.
\newblock {\slshape \bibinfo{journal}{CoRR}} \bibinfo{volume}{abs/2303.08774},
  \doi{10.48550/arXiv.2303.08774}.
\newblock \eprint{2303.08774}.

\bibitemdeclare{book}{russell2019human}
\bibitem{russell2019human}
\bibinfo{author}{Stuart \surnamestart Russell\surnameend}
  (\bibinfo{year}{2019}): \emph{\bibinfo{title}{Human compatible: Artificial
  intelligence and the problem of control}}.
\newblock \bibinfo{publisher}{Penguin}.

\bibitemdeclare{inproceedings}{waga2022dynamic}
\bibitem{waga2022dynamic}
\bibinfo{author}{Masaki \surnamestart Waga\surnameend},
  \bibinfo{author}{Ezequiel \surnamestart Castellano\surnameend},
  \bibinfo{author}{Sasinee \surnamestart Pruekprasert\surnameend},
  \bibinfo{author}{Stefan \surnamestart Klikovits\surnameend},
  \bibinfo{author}{Toru \surnamestart Takisaka\surnameend} \&
  \bibinfo{author}{Ichiro \surnamestart Hasuo\surnameend}
  (\bibinfo{year}{2022}): \emph{\bibinfo{title}{Dynamic shielding for
  reinforcement learning in black-box environments}}.
\newblock In: {\slshape \bibinfo{booktitle}{Automated Technology for
  Verification and Analysis: 20th International Symposium, ATVA 2022, Virtual
  Event, October 25--28, 2022, Proceedings}}, \bibinfo{organization}{Springer},
  pp. \bibinfo{pages}{25--41}.

\bibitemdeclare{article}{me}
\bibitem{me}
\bibinfo{author}{Francis~Rhys \surnamestart Ward\surnameend},
  \bibinfo{author}{Tom \surnamestart Everitt\surnameend},
  \bibinfo{author}{Francesca \surnamestart Toni\surnameend} \&
  \bibinfo{author}{Francesco \surnamestart Belardinelli\surnameend}
  (\bibinfo{year}{Forthcoming}): \emph{\bibinfo{title}{Honesty is the Best
  Policy: Defining and Mitigating AI Deception}}.

\end{thebibliography}
\end{document}